\documentclass[letterpaper, 10 pt, conference]{ieeeconf}  

\IEEEoverridecommandlockouts                              

\usepackage{graphicx}
\usepackage{cite}

\usepackage{pifont}

\usepackage{alltt}

\usepackage{enumerate}
\usepackage{siunitx}

\usepackage{epstopdf}
\usepackage{pbox}

\usepackage{multicol}

\usepackage{picinpar}
\usepackage{amsmath}
\usepackage{url}
\usepackage{flushend}
\usepackage{colortbl}
\usepackage{soul}
\usepackage{multirow}
\usepackage{amsfonts}
\usepackage{tabularx}
\usepackage{booktabs}
\usepackage{float}
\usepackage{cuted} 
\usepackage{caption} 

\usepackage{amssymb}  
\usepackage{subcaption}  
\usepackage[T1]{fontenc}
\usepackage{algorithm}
\usepackage{algorithmic}
\usepackage[hidelinks]{hyperref}

\usepackage{xcolor}

\usepackage{hyperref} 

\title{ Robust and Generalized Humanoid Motion Tracking}

\author{
\textnormal{Yubiao~Ma}$^{1}\dagger$,
\textnormal{Han~Yu}$^{2}\dagger$,
\textnormal{Jiayin~Xie}$^{2}$,
\textnormal{Changtai~Lv}$^{2}$,
\textnormal{Qiang~Luo}$^{2}$, 
\textnormal{Chi~Zhang}$^{2}$,\\
\textnormal{Yunpeng~Yin}$^{2}$,
\textnormal{Boyang~Xing}$^{2}$,
\textnormal{Xuemei~Ren}$^{1}$,
and~\textnormal{Dongdong~Zheng}$^{1,2}\ast$%
\thanks{$^{1}$Beijing Institute of Technology, Beijing, China.}%
\thanks{$^{2}$Humanoid Robotics (Shanghai) Co., Ltd., Shanghai 201203, China.}%
\thanks{$\dagger$These authors contributed equally.}%
\thanks{$\ast$Corresponding author: Dongdong Zheng (\texttt{ddzheng@bit.edu.cn}).}%
}

\begin{document}

\maketitle
\thispagestyle{empty}

\begin{strip}
    \vspace{-2.5cm}
    \centering
    \includegraphics[width=\textwidth]{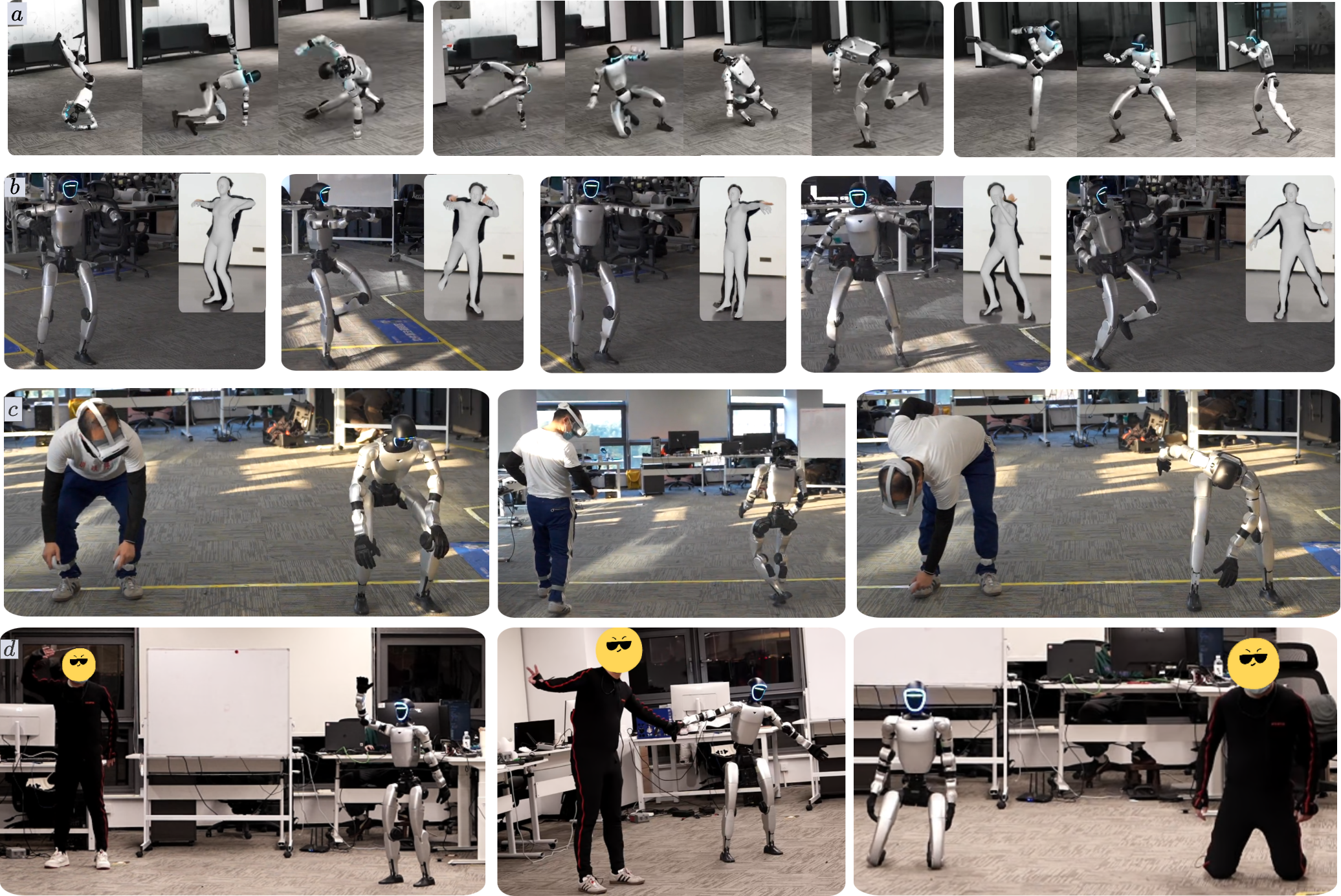}
    \captionof{figure}{Qualitative results illustrating the generalization of our method across different motion data sources, including public MoCap datasets, video-derived motions obtained from human pose estimation on public videos, and real-time teleoperation demonstrations via VR and a motion-capture suit.}
    \label{result}
\end{strip}

\begin{abstract}
Learning a general humanoid whole-body controller is challenging because practical reference motions can exhibit noise and inconsistencies
after being transferred to the robot domain, and local defects may be amplified by closed-loop execution,
causing drift or failure in highly dynamic and contact-rich behaviors.
We propose a dynamics-conditioned command aggregation framework that uses a causal temporal encoder to summarize recent proprioception
and a multi-head cross-attention command encoder to selectively aggregate a context window based on the current dynamics.
We further integrate a fall recovery curriculum with random unstable initialization
and an annealed upward assistance force to improve robustness and disturbance rejection.
The resulting policy requires only about $3.5$ hours of motion data and supports single-stage end-to-end training without distillation.
The proposed method is evaluated under diverse reference inputs and challenging motion regimes,
demonstrating zero-shot transfer to unseen motions as well as robust sim-to-real transfer on a physical humanoid robot.
\newline \textbf{Project page:} \url{https://zeonsunlightyu.github.io/RGMT.github.io/}

\end{abstract}

\section{Introduction}
Humanoid robots are compelling largely because of their generality.
Their morphology is naturally compatible with human environments, allowing them to operate within existing infrastructure and manipulate tools,
workspaces, and interfaces designed for people.
Moreover, their high-dimensional actuation and multi-contact capabilities support a broad spectrum of behaviors spanning locomotion, manipulation, and interaction~\cite{radosavovic2024nexttoken, radosavovic2024challengingterrain, WangH-RSS-25, Li-RSS-23, zhuang25a, HuangT-RSS-25, HeX-RSS-25, exbody, homie, falcon}.
To translate this vision into reliable operation,
it is necessary to develop a general whole-body controller that can coordinate the full body under changing contacts and task demands,
while maintaining stable behavior over long horizons.
However, this objective remains difficult to achieve in existing whole-body control research.
High-fidelity imitation in practice still often relies on training for a single motion or a small set of motions \cite{hub, kungfubot, asap,truong2025beyondmimic}, 
which tightly couples policy capability to a specific reference distribution and limits unified modeling and generalization across skills.

To learn diverse humanoid motions within a single policy, 
recent work has moved toward unified whole-body control that combines large-scale motion tracking objectives with broader motion coverage and more practical data acquisition. 
Several systems \cite{omnih2o,clone,twist2,exbody2} focus on scalable human to humanoid data collection and teleoperation, 
providing richer and more diverse demonstrations for training. 
In parallel, a number of motion tracking based controllers \cite{gmt,any2track,unitracker,kungfubot2} aim to directly train a universal tracker that can follow many motions under different disturbances. 
These approaches demonstrate promising progress toward general whole-body controllers on humanoid robots, 
but their tracking accuracy and closed loop stability remain suboptimal, 
especially during highly dynamic maneuvers and rapid contact transitions. 
SONIC \cite{sonic} further pushes motion coverage and naturalness, 
but it relies on very large-scale data and training resources, 
using more than 700 hours of motion data and substantial compute, 
which raises the barrier for iterative research and deployment.
Moreover, long-horizon operation in the real world requires not only stable tracking but also recovery after disturbances or falls, yet fall recovery is often not integrated into the main control policy \cite{HuangT-RSS-25, HeX-RSS-25}, limiting robustness and safety. 
Therefore, we need a general whole-body control framework that can learn effectively under limited data, maintain stable closed-loop tracking under imperfect references, and jointly integrate robustness improvements for highly dynamic and multi-contact scenarios with fall recovery within a single training and policy pipeline, enabling safe and continuous real-world task execution.


This paper presents a learning framework for general humanoid whole-body control.
The key idea is to condition the policy on the current dynamics, enabling it to interpret and aggregate contextual commands selectively rather than treating all reference signals as equally reliable supervision.
We obtain a compact dynamics representation from recent proprioceptive history using a causal temporal encoder and use it to guide command aggregation via multi-head attention \cite{attention}.
This design allows the policy to adaptively select and adjust reference segments under feasibility constraints imposed by the current dynamics, reducing the impact of noise and artifacts, particularly for highly dynamic motions and frequent contact transitions.
To further support safe and continuous real-world operation, we incorporate fall recovery into the same training framework \cite{HuangT-RSS-25, HeX-RSS-25}.
This integration broadens the experienced state distribution, improves robustness to disturbances, and strengthens tracking performance for contact-rich motion segments.
Our contributions are summarized as follows:
\begin{itemize}
  \item We propose a dynamics-conditioned command aggregation framework that combines causal temporal dynamics encoding with multi-head cross-attention.
  This design enables selective use of contextual commands under imperfect references and improves tracking accuracy and closed-loop stability in highly dynamic and contact-rich scenarios.
  The resulting general policy is trained end-to-end  using a compact dataset of about $3.5$ hours, without distillation or multi-stage training.
  \item We integrate fall recovery into a unified training framework. With randomized unstable initialization and an annealed external assistance force, a single policy learns stable control and self-recovery over a broader state distribution and contact conditions, leading to significantly improved robustness and disturbance rejection.
  \item We demonstrate strong generalization across diverse reference sources, including mocap, video-derived motions, and real-time full-body teleoperation.
  The learned policy transfers zero-shot to unseen motions and deploys robustly on the Unitree G1, enabling stable long-horizon tracking with integrated recovery and downstream applications such as joystick-driven locomotion.
\end{itemize}


\begin{figure*}[t]
  \centering
  \includegraphics[width=0.99\textwidth]{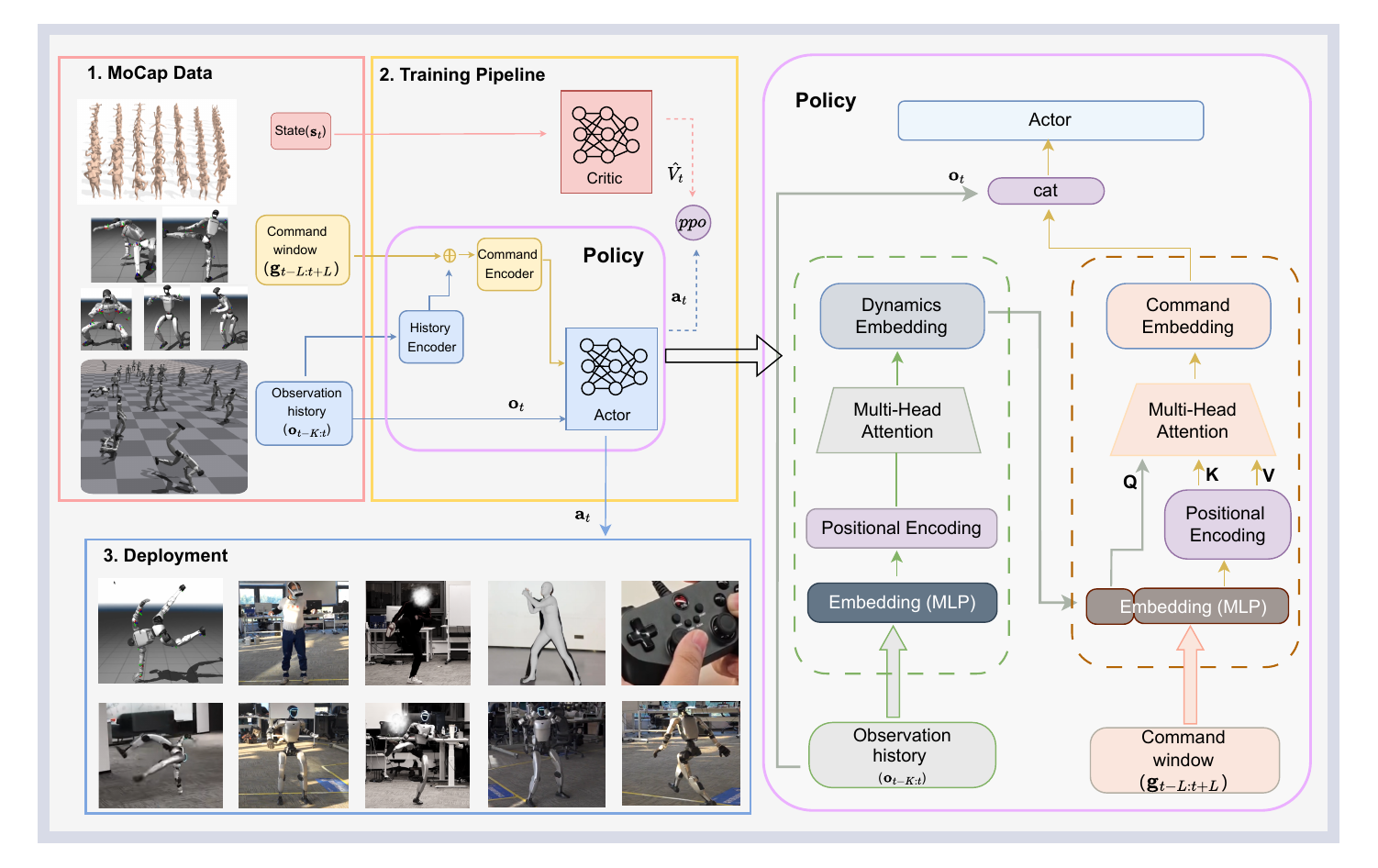}
  \caption{Overview of the proposed whole-body control pipeline. 
           A history encoder extracts a dynamics embedding from recent proprioception,
          which conditions a command encoder to aggregate the contextual command window.
          The resulting representation is fused with the current observation and fed to an actor-critic policy trained with PPO,
          and the learned actor is deployed for real-world whole-body motion tracking and teleoperation.}
  \label{architecture}
\end{figure*}

\section{METHOD}


\subsection{Humanoid Motion Dataset}
Our motion corpus is constructed from LAFAN1 \cite{lafan} and a selected subset of AMASS \cite{amass}, and all sequences are retargeted to our humanoid using General Motion Retargeting \cite{gmr}. 
In practice, large-scale mocap sources \cite{amass} and their retargeted counterparts often exhibit substantial redundancy and may include segments with low motion quality and inconsistent contacts. 
We therefore perform quality control to remove infeasible motions and low-quality sequences and to reduce redundancy, resulting in a relatively compact dataset of approximately $3.5$ hours. 

This quality-driven construction is crucial for general humanoid motion tracking. 
Even when the raw corpus is large, effective supervision can be limited by duplicated motions and low-quality clips, which can distract optimization and reduce training efficiency. 
In contrast, a smaller but diverse and higher-quality reference set provides cleaner and more informative supervision, improving generalization and closed-loop tracking accuracy. 
Importantly, enabled by our dynamics-conditioned command aggregation, this compact dataset is sufficient to train a strong general whole-body policy that is robust to noisy references and generalizes effectively to unseen motions.

\subsection{Motion Tracking Formulation}

\subsubsection{Observation Space}
Our policy receives an observation $\mathbf{o}_t$ that consists of a proprioceptive component and a command component.
The proprioceptive observation is
\begin{equation}
\mathbf{o}_t
=
\Big[
\mathbf{g}^{\mathrm{proj}}_t,\;
\boldsymbol{\omega}_t,\;
\mathbf{q}_t-\mathbf{q}_0,\;
\dot{\mathbf{q}}_t,\;
\mathbf{a}_{t-1}
\Big],
\end{equation}
where $\mathbf{g}^{\mathrm{proj}}_t\in\mathbb{R}^3$ denotes the gravity direction projected into the body frame, $\boldsymbol{\omega}_t\in\mathbb{R}^3$ is the base angular velocity, $\mathbf{q}_t\in\mathbb{R}^{29}$ and $\dot{\mathbf{q}}_t\in\mathbb{R}^{29}$ are joint positions and velocities, $\mathbf{q}_0$ is the default joint configuration, and $\mathbf{a}_{t-1}\in\mathbb{R}^{29}$ is the previous action.

The command observation provides per step reference targets extracted from the reference motion,
\begin{equation}
\mathbf{g}_t
=
\Big[
\mathbf{v}^{\mathrm{ref}}_t,\;
\boldsymbol{\omega}^{\mathrm{ref}}_t,\;
\mathbf{g}^{\mathrm{ref}}_t,\;
\mathbf{q}^{\mathrm{ref}}_t
\Big],
\end{equation}
where $\mathbf{v}^{\mathrm{ref}}_t\in\mathbb{R}^3$ and $\boldsymbol{\omega}^{\mathrm{ref}}_t\in\mathbb{R}^3$ are reference base linear and angular velocities expressed in the body frame, $\mathbf{g}^{\mathrm{ref}}_t\in\mathbb{R}^3$ is the reference gravity direction in the body frame, and $\mathbf{q}^{\mathrm{ref}}_t\in\mathbb{R}^{29}$ is the reference joint position at time $t$.

For value learning, we adopt an asymmetric actor-critic training scheme.
The critic additionally takes a privileged observation that facilitates more accurate value estimation,
\begin{equation}
\mathbf{o}^{\mathrm{priv}}_t
=
\Big[
h^{\mathrm{ref}}_t,\;
\mathbf{x}^{\mathrm{link}}_t,\;
\mathbf{v}_t
\Big],
\end{equation}
where $h^{\mathrm{ref}}_t$ denotes the reference base height, $\mathbf{x}^{\mathrm{link}}_t$ denotes body link poses, and $\mathbf{v}_t$ is the base linear velocity.
Accordingly, the critic input is
\begin{equation}
\mathbf{s}_t
=
\Big[
\mathbf{o}_t,\;
\mathbf{g}_t,\;
\mathbf{o}^{\mathrm{priv}}_t
\Big].
\end{equation}
All components of $\mathbf{s}_t$ used by the critic are noise-free, 
while the actor receives noisy observations $\mathbf{o}_t$.

\subsubsection{Action Space and Low-level Control}
The policy outputs a residual joint position command $\mathbf{a}_t \in \mathbb{R}^{29}$.
We interpret $\mathbf{a}_t$ as a corrective offset to the reference joint configuration and form the PD setpoint as
\begin{equation}
\mathbf{q}^{\mathrm{tar}}_t
=
\mathbf{q}^{\mathrm{ref}}_t
+
\mathbf{a}_t.
\end{equation}
Joint torques are computed by a joint-level PD controller,
\begin{equation}
\boldsymbol{\tau}_t
=
\mathbf{K}_p\Big(\mathbf{q}^{\mathrm{tar}}_t-\mathbf{q}_t\Big)
-
\mathbf{K}_d\,\dot{\mathbf{q}}_t,
\end{equation}
where $\mathbf{K}_p$ and $\mathbf{K}_d$ are diagonal gain matrices.

This residual formulation improves tracking accuracy by anchoring the PD setpoint to the reference motion while allowing  corrective adjustments.
It also makes exploration more efficient, since the policy searches around the reference pose rather than over the full joint configuration space.
As a result, training becomes more sample efficient and converges faster in practice.

\subsubsection{Reward Function}
Following \cite{truong2025beyondmimic}, 
we adopt a dense reward that combines reference tracking with safety and smoothness regularization. 
We define keypoints as a fixed set of links used in \cite{truong2025beyondmimic}, 
and formulate the tracking reward on these keypoints. 
Specifically, the tracking part measures keypoint alignment, relative pose consistency, 
and keypoint velocity consistency with exponential kernels. 
The regularization part penalizes rapid action changes, joint limit violations, 
and contacts on non-target body parts, which stabilizes learning and reduces physically implausible behaviors.

\subsection{Policy Learning Framework}

Multi-head attention (MHA) \cite{attention} provides an effective mechanism for query-conditioned information fusion. 
It computes content-based similarities between a query and a set of candidate features to produce adaptive aggregation weights, 
yielding a compact representation that emphasizes relevant elements while attenuating irrelevant or noisy ones. 
The multi-head formulation performs this matching in multiple learned subspaces, 
allowing the model to capture diverse relevance cues in parallel and increasing expressiveness beyond a single attention map. 
This capability is well suited to whole-body control, where command inputs are high dimensional and heterogeneous and may contain unreliable segments.

In practice, command sequences obtained from diverse sources often contain artifacts such as body penetration, 
inconsistent contacts, and high frequency noise. If the policy treats the entire command sequence as equally trustworthy supervision, 
abnormal segments can enter the representation with the same weight and be amplified in the action output, 
which can ultimately harm closed loop stability. Motivated by this observation,
we introduce dynamics-conditioned cross-attention in the policy architecture, as illustrated in Fig.~\ref{architecture}.
A dynamics representation is first extracted from recent proprioceptive history using causal temporal encoding and used as the query signal, 
and the contextual command window is then aggregated with adaptive attention weights. 
This design enables the policy to interpret and filter command information under physical feasibility constraints, 
reducing the influence of inconsistent reference signals on the control representation.

\subsubsection{History Encoder}
We encode the most recent $10$-step proprioceptive observations into a compact dynamics embedding.
The input sequence is
\begin{equation}
\mathbf{o}_{t-K:t}
=
\big[\mathbf{o}_{t-K},\ldots,\mathbf{o}_{t}\big],
\end{equation}
where each $\mathbf{o}_t\in\mathbb{R}^{93}$ and the sequence length is $K+1$.

Each observation is mapped to a token in an embedding space of dimension $n_{\mathrm{embd}}=128$ by a two-layer multi-layer perceptron (MLP),
\begin{equation}
\mathbf{E}_{t-K:t}
=
\mathrm{MLP}\!\left(\mathbf{o}_{t-K:t}\right)
\in \mathbb{R}^{(K+1)\times n_{\mathrm{embd}}},
\end{equation}
and a sinusoidal positional encoding is added to preserve temporal order,
\begin{equation}
\tilde{\mathbf{E}}_{t-K:t}
=
\mathbf{E}_{t-K:t}
+
\mathbf{P},
\qquad
\mathbf{P}\in\mathbb{R}^{(K+1)\times n_{\mathrm{embd}}}.
\end{equation}
The resulting token sequence is processed by a lightweight causal Transformer with multi-head self-attention.
Causality is enforced by a causal mask so that the token at time $\tau$ can attend only to tokens from times $\le \tau$ within the window.
Letting $\mathbf{H}^{(0)}=\tilde{\mathbf{E}}_{t-K:t}$, the causal self-attention block yields
\begin{equation}
\begin{aligned}
\mathbf{H}^{(1)} &= \mathbf{H}^{(0)}
+ \mathrm{MHA}\!\left(\mathrm{LN}\!\left(\mathbf{H}^{(0)}\right)\right),\\
\mathbf{H}^{(2)} &= \mathbf{H}^{(1)}
+ \mathrm{MLP}\!\left(\mathrm{LN}\!\left(\mathbf{H}^{(1)}\right)\right),\\
\bar{\mathbf{H}} &= \mathrm{LN}\!\left(\mathbf{H}^{(2)}\right),
\end{aligned}
\end{equation}
where $\mathrm{LN}(\cdot)$ denotes layer normalization.
Finally, we aggregate token features over time via element-wise max pooling to obtain the dynamics embedding $\mathbf{h}_t\in\mathbb{R}^{n_{\mathrm{embd}}}$:
\begin{equation}
\mathbf{h}_t[j]
=
\max_{\tau\in\{t-K,\ldots,t\}}
\bar{\mathbf{H}}_{\tau}[j],
\qquad
j=1,\ldots,n_{\mathrm{embd}}.
\end{equation}
This embedding extracts the robot dynamics from recent proprioceptive history and is used as the query signal for the subsequent dynamics-conditioned command encoder.

\subsubsection{Command Encoder}
The command encoder compresses a contextual command window into a compact latent representation while conditioning the aggregation on the current dynamics.
Its inputs are the dynamics embedding $\mathbf{h}_t\in\mathbb{R}^{n_{\mathrm{embd}}}$ and the command sequence
\begin{equation}
\mathbf{g}_{t-L:t+L}
=
\big[\mathbf{g}_{t-L},\ldots,\mathbf{g}_{t+L}\big],
\end{equation}
where each $\mathbf{g}_t\in\mathbb{R}^{38}$ and the window length is $2L + 1$.
The dynamics embedding is projected to the Transformer dimension through a two-layer multi-layer perceptron, yielding the query vector
\begin{equation}
\mathbf{q}_t
=
\mathrm{MLP}_{\mathrm{dyn}}\!\left(\mathbf{h}_t\right)
\in \mathbb{R}^{n_{\mathrm{embd}}}.
\end{equation}
In parallel, the command window is mapped to a token sequence in the same feature space using another two-layer multi-layer perceptron, and a sinusoidal positional encoding is added to preserve temporal order
\begin{equation}
\tilde{\mathbf{Z}}
=
\mathrm{MLP}_{\mathrm{cmd}}\!\left(\mathbf{g}_{t-L:t+L}\right)
+
\mathbf{P}^{\mathrm{cmd}},
\qquad
\tilde{\mathbf{Z}} \in \mathbb{R}^{(2L+1)\times n_{\mathrm{embd}}}.
\end{equation}
The encoder then applies a single dynamics-conditioned cross-attention block to aggregate the command tokens into a compact latent representation
\begin{equation}
\begin{aligned}
\mathbf{s}^{(1)}
&=
\mathbf{q}_t
+
\mathrm{MHA}\!\left(\mathrm{LN}\!\left(\mathbf{q}_t\right),\, \tilde{\mathbf{Z}}\right),\\
\mathbf{s}^{(2)}
&=
\mathbf{s}^{(1)}
+
\mathrm{MLP}\!\left(\mathrm{LN}\!\left(\mathbf{s}^{(1)}\right)\right),\\
\mathbf{u}_t
&=
\mathrm{LN}\!\left(\mathbf{s}^{(2)}\right)
\in \mathbb{R}^{n_{\mathrm{embd}}}.
\end{aligned}
\end{equation}
The resulting vector $\mathbf{u}_t$ serves as a compact, dynamics-conditioned command embedding at time $t$.
Because the cross-attention weights are conditioned on $\mathbf{h}_t$ through the query $\mathbf{q}_t$, the encoder adaptively emphasizes command elements that are more consistent with the current dynamics and down-weights unreliable segments, thereby reducing the influence of abnormal command artifacts on the control representation.

\begin{table*}[t]
  \centering
  \caption{Performance under different motion data sources.
  We report mean $\pm$ standard deviation.
  Higher is better for success rate, and lower is better for $E_{\mathrm{MPJPE}}$.}
  \label{tab:data}

  \renewcommand{\arraystretch}{1.15}
  \setlength{\tabcolsep}{5pt}

  \definecolor{blockgray}{RGB}{245,245,245}
  \definecolor{oursblue}{RGB}{234,242,255}
  \definecolor{ablationblue}{RGB}{242,247,255}
  \definecolor{ablationgreen}{RGB}{240,250,240}

  \sisetup{
    detect-weight = true,
    detect-family = true,
    uncertainty-mode = separate,
    uncertainty-separator = {\,\pm\,},
    table-number-alignment = center
  }

  \begin{tabularx}{0.98\textwidth}{@{}l
    >{\centering\arraybackslash}X >{\centering\arraybackslash}X
    >{\centering\arraybackslash}X >{\centering\arraybackslash}X
    >{\centering\arraybackslash}X >{\centering\arraybackslash}X@{}}
    \toprule

    \multirow{2}{*}{\textbf{Method}} &
    \multicolumn{2}{c}{\textbf{MoCap Data}} &
    \multicolumn{2}{c}{\textbf{Video-derived Motion}} &
    \multicolumn{2}{c}{\textbf{Ground-interaction Motion}} \\
    \cmidrule(lr){2-3} \cmidrule(lr){4-5} \cmidrule(lr){6-7}

    & \textbf{Succ.} $\uparrow$ & $\mathbf{E_{\mathrm{MPJPE}}}$ $\downarrow$
    & \textbf{Succ.} $\uparrow$ & $\mathbf{E_{\mathrm{MPJPE}}}$ $\downarrow$
    & \textbf{Succ.} $\uparrow$ & $\mathbf{E_{\mathrm{MPJPE}}}$ $\downarrow$ \\
    \midrule

    \rowcolor{blockgray}
    \multicolumn{7}{@{}l@{}}{\textbf{(a) Baseline}} \\

    GMT
    & 84.6\% & \num{65.15 +- 1.12}
    & 72.4\% & \num{96.47 +- 1.98}
    & 48.9\% & \num{146.95 +- 5.12} \\

    Any2Track
    & 89.2\% & \num{56.96 +- 0.91}
    & 54.3\% & \num{112.16 +- 3.96}
    & 41.2\% & \num{209.57 +- 4.22} \\

    {Ours}
    & \textbf{98.3\%} & \textbf{\num{41.12 +- 0.12}}
    & \textbf{94.6\%} & \textbf{\num{46.56 +- 0.28}}
    & \textbf{90.1\%} & \textbf{\num{54.92 +- 0.93}} \\

    \addlinespace[4pt]

    \rowcolor{ablationblue}
    \multicolumn{7}{@{}l@{}}{\textbf{(b) Ablations on Policy Architecture}} \\

    Ours SelfAttn CmdEnc
    & 89.8\% & \num{51.96 +- 0.72}
    & 76.7\% & \num{67.19 +- 1.31}
    & 73.2\% & \num{92.65 +- 2.31} \\

    Ours CNN HistEnc
    & 94.3\% & \num{48.61 +- 0.32}
    & 91.9\% & \num{53.13 +- 0.78}
    & 81.5\% & \num{68.92 +- 1.84} \\

    {Ours}
    & \textbf{98.3\%} & \textbf{\num{41.12 +- 0.12}}
    & \textbf{94.6\%} & \textbf{\num{46.56 +- 0.28}}
    & \textbf{90.1\%} & \textbf{\num{54.92 +- 0.93}} \\

    \addlinespace[4pt]

    \rowcolor{ablationgreen}
    \multicolumn{7}{@{}l@{}}{\textbf{(c) Ablations on Fall Recovery}} \\

    Ours w/o Fall Recovery
     & \textbf{98.4\%} & \textbf{\num{40.98 +- 0.09}}
    & \textbf{94.9\%} & \textbf{\num{46.16 +- 0.31}}
    & {70.5\%} & {\num{96.75 +- 2.77}} \\

    \addlinespace[4pt]

    {Ours}
    & {98.3\%} & {\num{41.12 +- 0.12}}
    & {94.6\%} & {\num{46.56 +- 0.28}}
    & \textbf{90.1\%} & \textbf{\num{54.92 +- 0.93}} \\

    \bottomrule
  \end{tabularx}
\end{table*}

\subsection{Fall Recovery Integration}

Automatic fall recovery is a key prerequisite for reliable humanoid deployment. 
Without an effective self-recovery mechanism, 
the system not only faces significant safety risks but also requires frequent human intervention, 
breaking task continuity. This issue is particularly pronounced for whole-body motion tracking, 
where rapid dynamics and frequent contact transitions can amplify closed-loop errors and trigger instability.
Therefore, we integrate a simple yet reliable automatic fall recovery mechanism into our whole-body control framework.

\subsubsection{Randomized Recovery Initialization}
We designate a subset of the parallel environments as recovery environments with probability $0.15$ 
and reset the robot in these environments to randomized poses, 
exposing the policy to a broad range of unstable configurations and contact initial conditions.
This process also enriches contact experience during training, 
since repeated falls and stand-ups naturally induce diverse ground-contact patterns, 
which improves tracking accuracy and generalization for motions with frequent or complex contact transitions.

For these recovery environments, we apply an upward pulling force with magnitude uniformly sampled from $[0, 200]$ to assist exploration 
at early training stages by increasing the frequency of recoverable states.
The assistance is linearly annealed over training iterations and reduced to a negligible level,
ensuring that the final policy performs fall recovery using only its own control.

\subsubsection{Termination Conditions}

We employ state-based episode termination and environment resets to maintain stable training and high-quality rollouts. 
For all environments, an episode is reset either when it reaches a predefined time limit or when an instability event is detected. 
Instability is identified by three conditions: excessive base orientation deviation, insufficient base height, and abnormally low height of key body links.

To learn automatic fall recovery, we introduce an additional termination strategy for the recovery environments.
Within a predetermined recovery window of $3$ seconds, recovery environments are not terminated early by the instability criteria, 
allowing the policy to complete stand-up and re-stabilization within the same episode.
If the robot fails to recover within this window, the episode is terminated and the environment is reset, 
avoiding prolonged rollouts in unrecoverable states and maintaining training efficiency.

\subsection{Training Setup}
We train our policy in the Isaac Gym simulator~\cite{isaacgym}. 
All training runs are conducted on a single NVIDIA RTX~4090 GPU with 5{,}680 parallel environments. 
We train on approximately 3.5 hours of motion data curated from subsets of LAFAN1~\cite{lafan} and AMASS~\cite{amass}, 
and observe strong generalization to previously unseen motions. 
The proprioceptive history length is set to $K=9$, and the command window half-length is set to $L=10$.

\section{Experiments}

In this section, we evaluate our approach on the 29 degrees-of-freedom (DOF) Unitree G1 \cite{unitree_g1} humanoid robot 
and demonstrate strong generalization and robustness in both simulation and the real world. 
We conduct quantitative comparisons against representative baselines and targeted ablation studies to validate improved whole-body motion tracking performance 
and robustness to noise in command inputs. 
Finally, we deploy the learned policy on the physical robot to showcase reliable tracking and generalization across diverse motions, 
and demonstrate its applicability to downstream tasks such as real-time teleoperation and online motion generation.

\begin{table}[t]
\centering
\caption{Noise specifications for command.}
\begin{tabularx}{0.49\textwidth}{|l|X|}
\hline
\textbf{Command} & \textbf{Noise Specification} \\
\hline
Base linear velocity jitter & $\Delta \mathbf{v}_t^{\mathrm{ref}} \sim \mathcal{U}([-0.5, 0.5]^3)$ (z: $\pm$0.2) \\
Base angular velocity jitter & $\Delta \omega^{\mathrm{ref}}_t \sim \mathcal{U}([-0.52, 0.52]^3)$  \\ 
Base gravity direction jitter & $\Delta \mathbf{g}_t^{\mathrm{ref}} \sim \mathcal{U}([-0.05, 0.05]^3)$ \\
Base joint position jitter & $\Delta \mathbf{q}_t^{\mathrm{ref}} \sim \mathcal{U}([-0.1, 0.1])$ \\
\hline
\end{tabularx}
\label{tab:noise}
\end{table}

\begin{figure*}[t]
  \centering
  \includegraphics[width=\textwidth]{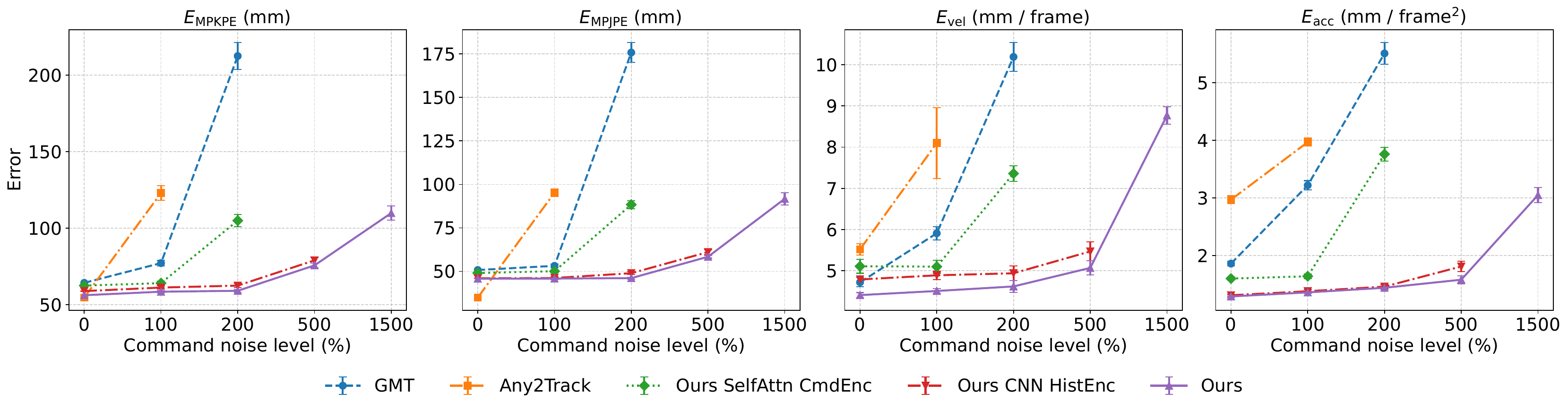}
  \caption{Robustness under reference command noise.}
  \label{fig:noise_robustness}
\end{figure*}

\subsection{Performance Evaluation}

\subsubsection{Comparison with Baselines}

We validate that the proposed approach learns a general motion tracker 
that achieves stable and accurate tracking over a wide range of previously unseen motions. 
We compare our method against two representative recent trackers, Any2Track~\cite{any2track} and GMT~\cite{gmt}, 
both of which provide officially released models. 
To ensure consistency and fairness, we directly use these models and evaluate all methods on the same datasets.
All methods are tested in MuJoCo~\cite{mujoco}, which is supported by the above baseline implementations and provides a unified evaluation platform. 
A comprehensive set of metrics is adopted to capture both pose accuracy and physical feasibility.
Specifically, the success rate (Succ) is defined as the fraction of rollouts without falls,
where a fall is identified when the root height deviates from the reference by more than $0.2$~m.
In addition, we report the mean per-joint position error (MPJPE) $E_{\mathrm{MPJPE}}$ (in mm),
which measures the 3D position error of joints and quantifies joint-level tracking accuracy.

Table~\ref{tab:data} (a) compares baseline performance across three motion sources: 
MoCap data curated from subsets of LAFAN1 and AMASS, video-derived motions, and ground-interaction motions. 
The MoCap subset comes from the same MoCap sources used to train all methods and thus measures in-distribution 
tracking performance. The video-derived motions are estimated from publicly available videos 
and cover common behaviors such as walking, dance, and martial arts. 
Since this subset is not used for training, it directly evaluates cross-source generalization. 
Across all three subsets, our method achieves the highest success rate and the lowest $E_{\mathrm{MPJPE}}$. 
The improvement is most pronounced on the video-derived subset, 
where we substantially outperform GMT and Any2Track, 
indicating stronger robustness to distribution shifts in the reference motions.

\subsubsection{Ablations on Policy Architecture}

Table~\ref{tab:data} (b) studies the effect of key design choices in our tracker. 
Replacing the causal history encoder with a CNN-based variant (\textit{Ours CNN HistEnc}) consistently degrades performance across all three data sources, 
indicating that the causal history encoder contributes to stable and accurate tracking. 
Replacing the  cross-attention command encoder with a self-attention variant (\textit{Ours SelfAttn CmdEnc}) leads to a much larger performance drop, 
particularly on the video-derived and ground-interaction subsets. This highlights the key role of dynamics-conditioned cross-attention in robust command aggregation under distribution shifts and command artifacts.

\subsubsection{Ablations on Fall Recovery}

Table~\ref{tab:data} (c) studies the effect of integrating fall recovery during training. 
Incorporating fall recovery does not noticeably compromise tracking accuracy or generalization to unseen motions, as reflected by comparable performance on the MoCap and video-derived subsets. 
In contrast, removing fall recovery training (\textit{Ours w/o Fall Recovery}) substantially degrades performance on the ground-interaction subset, which includes contact-intensive behaviors such as crawling, kneeling, sitting, and breakdance-style motions. 
In this setting, the success rate (Succ.) drops markedly and $E_{\mathrm{MPJPE}}$ increases. 
These results suggest that fall recovery not only provides self-recovery capability for safer long-horizon execution but also broadens contact experience during training, thereby improving robustness under complex ground interactions.

\subsection{Robustness Evaluation}

Robustness to command noise is essential for real-time teleoperation and generalization to unseen motions.
In practical applications, commands from human operators, motion estimation pipelines, or high-level planners are often noisy and subject to uncertainty.
To evaluate robustness under such conditions, we evaluate on the Charleston dance motion,
which is included in the training set of all methods and has been showcased by both GMT and Any2Track.
We inject varying levels of noise into the reference commands to systematically assess the robustness of each method.
The noise specifications (base noise patterns) are summarized in Table~\ref{tab:noise} .
We define the noise level (\%) by uniformly scaling the ranges in Table~\ref{tab:noise} .


In Fig.~\ref{fig:noise_robustness}, we report additional metrics beyond $E_{\mathrm{MPJPE}}$.
$E_{\mathrm{MPKPE}}$ measures the mean keypoint position error (in mm).
To assess physical fidelity, $E_{\mathrm{vel}}$ and $E_{\mathrm{acc}}$ measure the differences in keypoint velocities and accelerations relative to the reference motion,
reported in mm/frame and mm/frame$^{2}$, respectively.
As shown in Fig.~\ref{fig:noise_robustness}, baseline methods such as GMT and Any2Track degrade rapidly as the noise level increases.
$E_{\mathrm{MPJPE}}$, $E_{\mathrm{vel}}$, and $E_{\mathrm{acc}}$ rise substantially, and the baselines struggle to maintain stable tracking when the noise level exceeds 200\%.
These results suggest that, without an explicit mechanism to interpret and filter corrupted commands, baseline methods are more likely to amplify command noise in closed-loop execution, leading to instability at high noise levels.

In contrast, our method demonstrates significantly stronger robustness across all four metrics.
Even at high noise levels, the errors remain relatively low and degrade much more gradually.
Ablation experiments further confirm this trend.
Replacing the causal history encoder with a CNN-based variant consistently degrades robustness, confirming the importance of the causal history encoder.
Replacing the cross-attention command encoder with a self-attention variant leads to a much larger degradation, especially in the high-noise regime, highlighting the cross-attention module as the key mechanism for filtering noisy commands.
Notably, our full model maintains stable motion tracking even under noise levels up to 1500\%, demonstrating strong tolerance to severe command corruption.
Overall, these results highlight the central role of dynamics-conditioned cross-attention in filtering noisy commands and improving robustness, while the causal history encoder provides complementary benefits by stabilizing the dynamics representation used for command interpretation.

\subsection{Real-World Experiments}

\begin{figure}[t]
  \centering
  \includegraphics[width=0.48\textwidth]{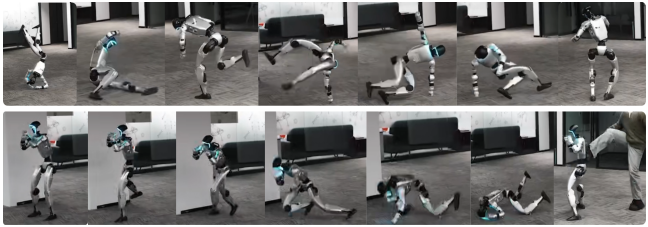}
  \caption{Real-world dance tracking with fall recovery.}
  \label{fig:play}
\end{figure}


\subsubsection{Robust Whole-Body Motion Tracking}
Fig.~\ref{fig:play} qualitatively illustrates two challenging behaviors enabled by our policy.
In the top row, the robot tracks a breakdance-style motion with frequent ground contacts and rapid contact transitions, demonstrating our policy's ability to robustly coordinate whole-body motion under complex contact patterns.
In the bottom row, we apply an external push that causes the robot to fall. The policy autonomously executes a recovery maneuver and then resumes the motion-tracking task without manual resets, which improves robustness for long-horizon deployment and reduces the need for human intervention.

\subsubsection{Video-Derived Motion Tracking}
Fig.~\ref{result}(b) presents qualitative results on video-derived motions to evaluate generalization to unseen motions.
For these examples, the reference commands are produced by a video-based human motion estimation pipeline~\cite{shen2024gvhmr}, which introduces noise and distribution shift.
Despite these challenges, our policy tracks the commanded motions on the real robot with high fidelity, reproducing the overall timing and whole-body coordination.
These results demonstrate effective transfer to previously unseen motion styles.

\subsubsection{Real-Time Teleoperation}
We develop a real-time full-body teleoperation pipeline with two different motion-tracking front ends: a consumer-grade PICO VR whole-body tracking interface and a wearable motion-capture suit.
For the PICO setup, the operator wears a VR headset, two ankle trackers, and handheld controllers, and the system outputs full-body pose estimates at runtime.
For the motion-capture suit setup, the suit streams full-body pose measurements online through its vendor SDK, which are converted to the same kinematic representation to ensure a unified downstream interface. 
In both cases, the estimated human motion is streamed in real-time and transformed into our reference command representation $\mathbf{g}_t$, which is then consumed by the deployed actor together with proprioceptive observations. 
This teleoperation setting is substantially noisier than offline MoCap due to sensing drift, latency, and operator inconsistency, and thus serves as a practical stress test of generalization and robustness. 
As shown in Fig.~\ref{result}(c) and~(d), under both teleoperation sources, our policy remains stable and precisely tracks the incoming commands without manual resets, enabling challenging whole-body behaviors such as crawling, high kicks, and deep squats.

\subsubsection{Joystick-driven locomotion}
\begin{figure}[t]
  \centering
  \includegraphics[width=0.48\textwidth]{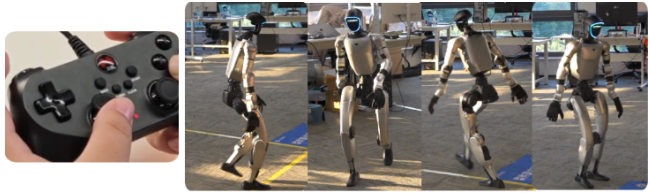}
  \caption{Joystick-driven stylized locomotion.}
  \label{fig:joystick}
\end{figure}

Fig.~\ref{fig:joystick} illustrates a downstream integration of our tracker into a representative computer graphics motion synthesis method~\cite{learnedmotionmatching}.
A handheld game joystick provides high-level locomotion commands, which are mapped to a sequence of reference motion targets and streamed to our policy as contextual commands.
Although the synthesized reference sequence can exhibit abrupt transitions due to discrete clip-switching and matching artifacts, our policy remains stable and tracks the commanded base velocity in a stylized manner, producing coherent whole-body locomotion on the real robot.
This result suggests that our tracker can serve as a robust low-level controller for upstream motion generation modules, tolerating non-smooth reference trajectories while preserving responsive velocity tracking and motion style.

\section{Conclusion} 
\label{sec:conclusion}

We propose a whole-body motion tracking framework that achieves robust, generalized humanoid control with a single policy.
By conditioning command aggregation on a dynamics representation extracted from recent proprioception, the policy can down-weight inconsistent reference segments and remain stable under substantial command corruption.
Results in simulation and on a 29-DoF Unitree G1 demonstrate accurate tracking across diverse motion sources, strong generalization to previously unseen motions, and reliable execution under complex contact patterns and external disturbances, enabling practical applications such as real-time teleoperation and joystick-driven locomotion.

Future work will incorporate global localization to enable long-horizon, world-frame consistent tracking, and expand integration with upstream motion generation and planning modules to support richer downstream tasks.


\bibliographystyle{IEEEtran}
\bibliography{IEEEabrv,main}  

\end{document}